# Decision-Theoretic Planning with Concurrent Temporally Extended Actions

**Khashayar Rohanimanesh**
Department of Computer Science
Michigan State University
East Lansing, MI 48824-1226
khash@cse.msu.edu

**Sridhar Mahadevan**
Department of Computer Science
Michigan State University
East Lansing, MI 48824-1226
mahadeva@cse.msu.edu

## Abstract

We investigate a model for planning under uncertainty with temporally extended actions, where multiple actions can be taken concurrently at each decision epoch. Our model is based on the *options* framework, and combines it with factored state space models, where the set of options can be partitioned into classes that affect disjoint state variables. We show that the set of decision epochs for concurrent options defines a semi-Markov decision process, if the underlying temporally extended actions being parallelized are restricted to Markov options. This property allows us to use SMDP algorithms for computing the value function over concurrent options. The concurrent options model allows overlapping execution of options in order to achieve higher performance or in order to perform a complex task. We describe a simple experiment using a navigation task which illustrates how concurrent options results in a more optimal plan when compared to the case when only one option is taken at a time.

## 1 Introduction

In our everyday life, our brain is constantly planning and executing concurrent (parallel) behaviors. For example, when we are driving, in parallel we visually search for road signs, while we may be talking to a passenger. Or when walking toward our car in a parking lot or our office, we may simultaneously reach for our keys, while continuing to talk on a cell-phone and navigating through the environment. Parallel execution of behaviors is sometimes useful in performing a task more quickly (e.g., the parking lot example). In other situations, the nature of the task requires that multiple behaviors run concurrently and cooperatively in order to perform the task (in the driving example, we have to both look at the road and navigate the car simultaneously). In this paper, we investigate a model for planning with concurrent behaviors. We adopt the theoretical framework of *options* (Sutton et al., 1999) to model temporally extended actions, since it is both a well-developed rigorous framework that addresses planning under uncertainty with temporally extended actions, and it allows *looking inside* behaviors to improve composition of temporally extended actions.

Previous work on decision-theoretic concurrent planning seems to be largely restricted to work focusing on combining *primitive* actions, ranging from planning in multi-dimensional vector action spaces (Cichosz, 1995) to planning with multiple simultaneous MDPs (Singh & Cohn, 1998), where the composite state space is the cross product of the state spaces of each individual MDP and the action set is a proper subset of a multi-dimensional primitive action space. In these models, each decision epoch is fixed and equal to single step execution. Our work differs in that we address planning with a set of parallel temporally extended actions that may not terminate at the same time, which makes the problem more challenging. We also exploit the fact that in many real world problems, the set of options can be factored into those that affect disjoint state variables. This factoring greatly reduces the complexity of planning with multi-dimensional composite state and action spaces (Boutilier et al., to appear).

In this paper, we address planning with a set of concurrent options, assuming that they do not compete for a shared resource (in the parking lot example, the option of reaching for the car key and the option of walking affect different portions of the composite state space). We present a navigation task involving moving through rooms using keys to open locked doors to illustrate how the concurrent options model facilitates faster planning. Our experiments show that the concurrent options model improves performance com-pared to the sequential case when only one behavior at a time can be executed.

The rest of the paper is organized as follows. In section 2, we will briefly overview the option framework. In section 3 we define the concurrent options model in detail. In section 4 we will present a computational problem and the performance results of planning using the concurrent options model. Section 5 outlines some problems for future research.

## 2　Options

Options are a generalization of primitive actions that include temporally extended courses of action in the context of reinforcement learning (Sutton et al., 1999). Options consist of three components: a policy $\pi : S \times A \to [0,1]$, a termination condition $\beta : S \to [0,1]$, and an initiation set $I \subseteq S$, where $I$ denotes the set of states $s \in S$ in which the option can be initiated. Note that we can restrict the scope of application of a particular option by controlling the initiation set and the termination condition. For any state $s$, if option $\pi$ is taken, then primitive actions are selected based on $\pi$ until it terminates according to $\beta$. An option $O$ is a *Markov option* if its policy, initiation set and termination condition depend stochastically only on the current state $s \in S$. Given a set of options $O$, let $O_s$ denote the set of options in $O$ that are available in each state $s \in S$ according to their initiation set. $O_s$ resembles $A_s$ in the standard reinforcement learning framework, in which $A_s$ denotes the set of primitive (single step) actions. Similarly, we introduce *policies over options*. For a decision epoch $d_t$, the Markov policy over options $\mu : S \times O \to [0,1]$ selects an option $o_t \in O$, according to the probability distribution $\mu(s_t, .)$. The option $o_t$ is then initiated in $s_t$ until it terminates at a random time $t + k$ in some state $s_{t+k}$ according to the termination condition, and the process repeats in $s_{t+k}$. For an option $o \in O$, and for any state $s \in S$, let $\varepsilon(o, s, t)$ denote the event of $o$ being initiated in state $s$ at time t. The total discounted reward accrued by executing option $o$ in any state $s \in S$ is defined as:

$$r_s^o = E\{r_{t+1} + \gamma r_{t+2} + ... + \gamma^{k-1} r_{t+k} \mid \varepsilon(o, s, t)\}$$

where $t + k$ is the random time at which $o$ terminates. Also let $p^o(s, s', k)$ denote the probability that the option $o$ is initiated in state $s$ and terminates in state $s'$ after $k$ steps. Then

$$p_{ss'}^o = \sum_{k=1}^{\infty} p^o(s, s', k) \gamma^k$$

Given the reward and state transition model of option $o$, we can write the Bellman equation for the value of a general policy $\mu$ as:

$$V^\mu(s) = \sum_{o \in O_s} \mu(s, o) \left[ r_s^o + \sum_{s'} p_{ss'}^o V^\mu(s') \right]$$

Similarly we can write the "option-value" Bellman equation for the value of an option $o$ in state $s$ as

$$Q^\mu(s, o) = r_s^o + \sum_{s'} p_{ss'}^o \sum_{o' \in O_{s'}} \mu(s', o') Q^\mu(s', o')$$

and the corresponding *optimal* Bellman equations are as follows:

$$V_O^*(s) = \max_{o \in O_s} \left[ r_s^o + \sum_{s'} p_{ss'}^o V_O^*(s') \right]$$

$$Q_O^*(s, o) = r_s^o + \sum_{s'} p_{ss'}^o \max_{o' \in O_{s_{t+k}}} Q^*(s', o')$$

We can use *synchronous value iteration* (SVI) to compute $V_O^*(s)$ and $Q_O^*(s, o)$, which iterates the following step for every state $s \in S$:

$$V_t(s) = \max_{o \in O_s} \left[ r_s^o + \sum_{s'} p_{ss'}^o V_{t-1}(s') \right]$$

$$Q_t(s, o) = r_s^o + \sum_{s'} p_{ss'}^o \max_{o' \in O_{s'}} Q_{t-1}(s', o')$$

Alternatively, if the option model is unknown, we can estimate $Q_O^*(s, o)$ using SMDP *Q-learning*, by doing sample backups after the termination of each option $o$, which transitions from state $s$ to $s'$ in $k$ steps with cumulative discounted reward $r$:

$$Q(s, o) \leftarrow Q(s, o) +$$
$$\alpha \left[ r + \gamma^k \max_{o' \in O_{s'}} Q(s', o') - Q(s, o) \right]$$

## 3　Concurrent Options

**Definition:** Let $o \equiv <I, \pi, \beta>$ be an option with state space $S_o$ governed by the set of state variables $W_o = \{w_1^o, w_2^o, ..., w_{n_o}^o\}$. Let $\varphi_o \subset W_o$ denote the subset of state variables that evolve by some other processes (e.g. other options) and independent of $o$, and let $\Omega_o = W_o - \varphi_o$ denote the subset of state variables that evolve solely based on the option $o$. There is no explicit restriction on the initiation set, policy and termination condition with respect to the state space $S_o$. We refer to the class of options with this property as *partially-factored* options. As an example, consider the task of delivering parts to a set of machines in a factory environment. The agent has to load up a part from the inventory load station and deliver it to a particular machine. We may define the options *deliver-part* and *load-part* with the set of state variables $W =$ $W_{deliver\_part} = W_{load\_part} = \{position, part\_ready\}$ denoting the position of the agent and whether or not a part is available at the inventory load station, respectively. We may also define an *inventory* option with state variable $W_{inventory} = \{part\_ready\}$, which is set or reset whenever a part is ready to load, or not available. It is clear from this example that even though the initiation set, policy and the termination condition of the options *deliver-part* and *load-part* are defined over the whole state space spanned by $W$, the execution of these options has no effect on the state variable *part-ready* that is controlled by some other option (e.g. *inventory* option). In this example $\varphi_{deliver\_part} = \varphi_{load\_part} = \{part\_ready\}$, and $\Omega_{deliver\_part} = \Omega_{load\_part} = \{position\}$. Also $\varphi_{inventory} = \emptyset$ and $\Omega_{inventory} = \{part\_ready\}$. Two options $o_i$ and $o_j$ are called *coherent* if (1) they are both *partially-factored* options, and (2) $\Omega_{o_i} \cap \Omega_{o_j} = \emptyset$ (this condition is required to ensure that these two options will not affect the same portion of the state space so that they can safely run in parallel). In the above example, *deliver-part* and *inventory* options are coherent, but *deliver-part* and *load-part* options are not coherent, since the state variable *position* is controlled by both *deliver-part* and *load-part* options. Now, assume $O$ is a set of available options and $\{C_1, C_2, ..., C_n\}$ are $n$ classes of options that partition $O$ into $n$ disjoint classes such that any two options belonging to different classes are coherent (can run in parallel), and any two options within the same class are not coherent (they control shared state variables and cannot run in parallel). Clearly any set of options generated by drawing each option from a separate class can safely be run in parallel (for the above example, we can define two classes of options with this property: $C_1 = \{deliver\_part, load\_part\}$ and $C_2 = \{inventory\}$).

Given the above definitions, we can define the *Concurrent Options* model as a 4-tuple (S, A, P, R):

**State space:** The state space represented by $S$ is spanned by the set of state variables in the union of $W_o$ sets ($o \in O$):

$$S = \bigcup_{o \in O} S_o \quad \text{and} \quad W = \bigcup_{o \in O} W_o \quad (8)$$

It is also simple to verify that:

$$W = \bigcup_{o \in O} \Omega_o \quad (9)$$

Let $S_{\Omega_o}$ denote the sub-space in $S_o$ that is spanned by $\Omega_o$. Note that for every option $o$, $S_{\Omega_o}$ is a sub-space of $S_o$ and $S_o$ is a sub-space of $S$. For every option $o \in O$ let $\theta_{\Omega_o}(s_o)$ return a vector whose elements are the current value of each state variable in $\Omega_o$ for a given state $s_o \in S_o$, and let $\theta_{\Omega_o}(s)$ return a vector whose elements are the current value of each state variable in $\Omega_o$ for a given state $s \in S$. Also, let $\theta_o(s)$ return the current $s_o \in S_o$. It is simple to verify that

$$\forall o \in O, \ \theta_{\Omega_o}(s_o) = \theta_{\Omega_o}(s) \quad (10)$$

Based on equation 9, a sample state vector $s \in S$ can be represented by

$$s = (\theta_{\Omega_{o_1}}(s) : \theta_{\Omega_{o_2}}(s) : ... : \theta_{\Omega_{o_n}}(s)) \quad (11)$$

where ':' is the concatenation operator. We will use this notation to explain the other components of the model.

**Actions:** For every state $s \in S$, a set of one or more options, each belonging to a different class can be initiated concurrently, therefore:

$$\forall s \in S, \vec{O}(s) \subseteq C_1 \times C_2 \times ... \times C_n \quad (12)$$

A concurrent option $\vec{o} \in \vec{O}(s)$ consists of a set of $m$ ($1 \leq m \leq n$) options that can be initiated in parallel. We represent a concurrent option by $\vec{o} = (o_1, o_2, ..., o_m)$ and call it a *multi-option* (to distinguish them from regular options), and also represent the set of available options in state $s$ by $\vec{O}(s)$. We also need to define the event of termination of a multi-option $\vec{o}$. When multi-option $\vec{o}$ is executed in state $s$, a set of $m$ options $o_i \in \vec{o}$ are initiated. Each option $o_i$ will terminate at some random time $t_{o_i}$. We can define the event of termination for a multi-option based on either of the following events: (1) when any of the options $o_i \in \vec{o}$ terminates according to $\beta_i(s)$, multi-option $\vec{o}$ is declared terminated and the rest of the options that are not terminated at that point in time, are interrupted, or (2) when all of the options are terminated. From now on, we use $T1$ to refer to the first definition of the termination event, and $T2$ to refer to the second definition of the termination event. Mathematically, $T1$ and $T2$ can be defined as:

$$T1 \equiv min\{t_{o_i} \mid 1 \leq i \leq m\}$$
$$T2 \equiv max\{t_{o_i} \mid 1 \leq i \leq m\}$$

**Transition Probabilities:** We now define the state transition probabilities based on the termination event explained above. Let $P_{ss'}^{\vec{o}}$ denote the probability that multi-option $\vec{o} = (o_1, o_2, ..., o_m)$ is initiated in state $s$, and terminates in state $s'$. Based on $T1$, it denotes the probability that every option $o_i \in \vec{o}$ is initiated in state $s$, and when at least one of them terminates (according to its $\beta_i$ function), the state of the system is $s'$. Based on $T2$, it denotes the probability that every option $o_i \in \vec{o}$ is initiated in state $s$, and when all of them terminate (according to its $\beta_i$ function), the state of the system is $s'$. For each option $o_i$, the transition probability when only that option is initiated is well defined and is part of the option model. When more than one option is initiated simultaneously, the transition probability of the multi-option $\vec{o}$ is computed as follows. Let $P^{\vec{o}}(s, s', k)$ denote the the probability that multi-option $\vec{o}$ is initiated in state $s$, and terminates in state $s'$ after $k$ steps. Then:

$$P^{\vec{o}}_{ss'} = \sum_{k=1}^{\infty} P^{\vec{o}}(s, s', k) \gamma^k \quad (13)$$

Now, let $\xi^{\vec{o}}(s, s', k)$ denote the probability that multi-option $\vec{o}$ is initiated in state $s$, and after $k$ steps transitions to state $s'$. We can compute $P^{\vec{o}}(s, s', k)$ based on $\xi^{\vec{o}}(s, s', k)$, for either of the termination events explained above:

**Termination condition I (T1):** It can be viewed as the probability of initiating the multi-option $\vec{o}$ in state $s$, running it for $k$ steps without any of the options being terminated until step $k$, and at least one of the options terminates at step $k$:

$$P^{\vec{o}}(s, s', k) = \xi^{\vec{o}}(s, s', k) \left(1 - \prod_{i=1}^{m}(1 - \beta_i(s'))\right) \quad (14)$$

The second term on the right hand side in equation 14 denotes the probability that at least one of the options terminates in state $s'$ according to its termination condition $\beta_i$.

**Termination condition II (T2):** For this case, for any $k$, we have to include every possible permutation of termination of $m$ options $\{o_1, o_2, ..., o_m\}$ within $k$ steps. It can be computed based on the following recurrent equation:

$$P^{\vec{o}}(s, s', k) = \sum_{i=1}^{k} \sum_{\vec{o}' \subset \vec{o}} \sum_{s_u \in S} \begin{pmatrix} \xi^{\vec{o}}(s, s_u, i) \\ \times \\ \prod_{o \in \vec{o}'} \beta_o(s_u) \\ \times \\ P^{\vec{o}-\vec{o}'}(s_u, s', k-i) \end{pmatrix} \quad (15)$$

The first summation runs over all intermediate steps $i \leq k$ at which no termination happens. The second summation comprises every subset $\vec{o}'$ of the multi-option $\vec{o}$ that terminates at step $i$, and the third summation runs over every possible intermediate state in which $\vec{o}'$ terminates in $i \leq k$ steps. The first term inside the summation denotes the transition probability that the multi-option $\vec{o}$ initiated in state $s$ transitions to the intermediate state $s_u$ after $i$ steps; the second term denotes the probability that all of the options $o \in \vec{o}'$ (where $\vec{o}'$ is a subset of the original multi-option $\vec{o}$) terminate at the intermediate state $s_u$; and the third term denotes the $k - i$ steps transition probability of the rest of the options $(\vec{o} - \vec{o}')$. The following equation denotes the stopping criterion for the recurrent equation 15:

$$P^{\vec{o}}(s, s', 1) = \xi^{\vec{o}}(s, s', 1) \prod_{o \in \vec{o}} \beta_o(s') \quad (16)$$

in which $\xi^{\vec{o}}(s, s', 1)$ denotes the single step transition probability. Now, for every option $o_i$, $\xi^{o_i}(s_i, s'_i, 1)$ (where $s_i, s'_i \in S_i$) denotes the single step transition probability that the option $o_i$ is executed in state $s_i$ for only one step and the next state is $s'$, assuming that no other option is initiated. Since each option $o_i$ affects the state space through the set of state variables in $\Omega_{o_i}$ (and $\Omega_{o_i}$ sets are disjoint), we can define the $\xi^{o_i}(s_i, s'_i, 1)$ in terms of $\Omega_{o_i}$. Using equation 10 we get:

$$\xi^{o_i}(s_i, s'_i, 1) = \xi^{o_i}(\theta_{\Omega_{o_i}}(s), \theta_{\Omega_{o_i}}(s'), 1) \quad (17)$$

Using equation 11, we can rewrite the single step transition probability[1] of a multi-option $\vec{o}$:

$$\xi^{\vec{o}}(s, s', 1) = P^{\vec{o}}((\theta_{\Omega_{o_1}}(s) : \theta_{\Omega_{o_2}}(s) : ... : \theta_{\Omega_{o_m}}(s)),$$
$$(\theta_{\Omega_{o_1}}(s') : \theta_{\Omega_{o_2}}(s') : ... : \theta_{\Omega_{o_m}}(s')),$$
$$1) \quad (18)$$

Thus, using equation 17 we get:

$$\xi^{\vec{o}}(s, s', 1) = \prod_{i=1}^{m} \xi^{o_i}(\theta_{\Omega_{o_i}}(s), \theta_{\Omega_{o_i}}(s'), 1)$$
$$= \prod_{i=1}^{m} \xi^{o_i}(s_i, s'_i, 1) \quad (19)$$

Therefore, the single step transition probability of a multi-option $\vec{o}$ from state $s$ to state $s'$ is the product of single step transition probabilities of each individual option $o \in \vec{o}$ from state $s$ to state $s'$, since through single step execution of every option $o \in \vec{o}$ in state $s$, each option will control a disjoint set of state variables that mutually contributes to the formation of state $s'$ through their $\Omega$ sets, which is independent of the single-step execution of the other options that are running in parallel. Having defined the single step transition probability (equation 19), we can recursively define the k-step transition probability:

$$\xi^{\vec{o}}(s, s', k) = \sum_{s_j \in S} \begin{pmatrix} \xi^{\vec{o}}(s, s_j, k-1) \\ \times \\ \prod_{i=1}^{m}(1 - \beta_i(s_j)) \\ \times \\ \xi^{\vec{o}}(s_j, s', 1) \end{pmatrix} \quad (20)$$

---

[1]Note that in this representation, for clarity, we have omitted those state variables that are not covered by the union of $\Omega_o$ sets ($\forall o \in \vec{o}$). In such cases, we can explicitly list those state variables, since they remain unchanged.

in which the first term on the right hand side denotes the probability of executing multi-option $\vec{o}$ (initiated in state $s$) for $k-1$ steps and ending up in state $s_j$, the second term denotes the probability that none of the options $o_i \in \vec{o}$ terminate at state $s_j$ and the last term denotes the probability that the option is initiated in state $s_j$ and executed for a single step, and ended up in state $s'$. Based on equations 13, 14, 15, 17, 19 and 20, we can compute $P^{\vec{o}}_{ss'}$, for all $s, s' \in S$ and for all $\vec{o} \in \vec{O}(s)$.

**Reward function**: For any state $s \in S$ and for any multi-option $\vec{o} \in \vec{O}(s)$, it can be defined as:

$$R^{\vec{o}}_s = E\{r_{t+1} + \gamma r_{t+2} + \ldots + \gamma^{k-1} r_{t+k} \mid \varepsilon(\vec{o}, s, t)\} \quad (21)$$

where $t + k$ is the random time at which multi-option $\vec{o}$ terminates.

**State value function**: For any Markov policy $\mu$, the state value function can be written

$$V^\mu(s) = E\{r_{t+1} + \gamma r_{t+2} + \ldots + \gamma^{k-1} r_{t+k} + \gamma^k V^\mu(s_{t+k}) \mid \varepsilon(\mu, s, t)\}$$
$$= \sum_{\vec{o} \in \vec{O}(s)} \mu(\vec{o}, s) [R^{\vec{o}}_s + \sum_{s' \in S} P^{\vec{o}}_{ss'} V^\mu(s')]$$

where $k$ is the duration of multi-option $\vec{o}$ according to the termination event explained above.

**Multi-option value of $\vec{o}$ under Markov policy $\mu$**:

$$Q^\mu(s, \vec{o}) = E\{r_{t+1} + \gamma r_{t+2} + \ldots + \gamma^{k-1} r_{t+k} + \gamma^k V^\mu(s_{t+k}) \mid \varepsilon(\vec{o}, s, t)\}$$
$$= E\{r_{t+1} + \gamma r_{t+2} + \ldots + \gamma^{k-1} r_{t+k} + \gamma^k \sum_{\vec{o}' \in \vec{O}(s_{t+k})} \mu(s_{t+k}, \vec{o}') Q^\mu(s_{t+k}, \vec{o}') \mid \varepsilon(\vec{o}, s, t)\}$$
$$= R^{\vec{o}}_s + \sum_{s' \in S} P^{\vec{o}}_{ss'} \sum_{\vec{o}' \in \vec{O}(s')} \mu(s', \vec{o}') Q^\mu(s', \vec{o}')$$

where $k$ is the duration of multi-option $\vec{o}$ according to the termination condition explained above. It has been shown earlier that the set of Markov options defines a semi-Markov decision process (SMDP) (Sutton et al., 1999). It is natural to conjecture whether this result carries over to multi-options. We show that this is indeed the case, with the assumptions discussed above.

**Theorem (MDP + Concurrent Options = SMDP)**: For any MDP, and any set of *concurrent Markov options* defined on that MDP, the decision process that selects only among multi-options, and executes each one until its termination according to the multi-option termination condition, forms a semi-Markov decision process.

**Proof**: (Sketch) For a decision process to be a SMDP, it is required to define (1) set of states, (2) set of actions, (3) an expected cumulative discounted reward defined for every pair of state and action and (4) a well defined joint distribution of the next state and next decision epoch. In the concurrent options model, we have defined the set of states and the set of actions are the multi-options. The expected cumulative discounted reward and joint distributions of the next state and next decision epoch have been defined in terms of the underlying MDP. The policy and termination condition for every option that belongs to a multi-option, and the termination condition for a multi-option have also been defined.

## 4 Experimental Results

In this section we present a simple computational example that illustrates planning with concurrent options. We adopt the *rooms example* from (Sutton et al., 1999) and we add doors in each of the four hallways (Figure 1).

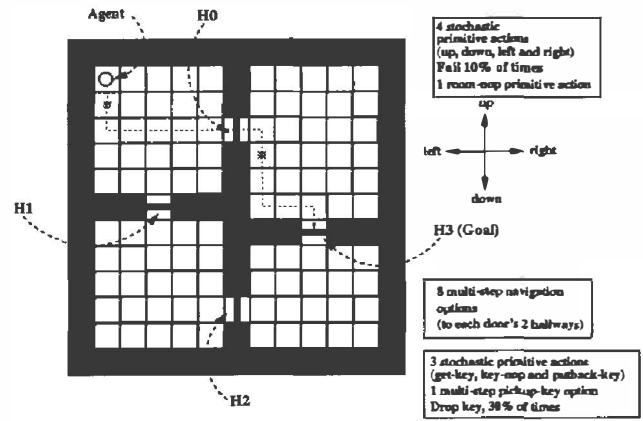

Figure 1: The rooms example with locked doors.

The agent cannot pass through locked (closed) doors unless it is holding the key. The state of the environment for this example consists of three state variables: position of the agent in the environment (represented by cells), state of the doors, and the state of the key. At any state, the agent can select actions from the set of navigation actions or the set of key related actions. Navigation actions comprises four stochastic primitive actions: *up*, *down*, *left* and *right* (Figure 1). Each navigation action with probability 9/10 causes the agent to move one cell in the corresponding direction, and with probability 1/10, moves the agent in one of the other three directions, each with probability 1/30. In either case, if the movement would take the agent into a wall, or a closed door when the agent is not holding the key, then the agent will remain in the same cell. We have also defined a *room-nop* primitive action that does not change the position of the agent (with probability 1). In each of the four rooms, we define two hallway Markov options (multi-step) that take the agent from anywhere within the room to one of the two hallway cells leading out of the room.

Figure 2 shows the policy for one of the hallway options. The termination condition $\beta(.)$ for hallway options is zero for states inside the room, except for the cell next to the target hallway cell (shaded cell in the Figure 2), in which the termination condition also depends on the state of the door in the target hallway, and also whether or not the agent is holding the key. In this cell, the termination condition is zero if either the door is open, or the door is closed and the agent is holding the key, otherwise the hallway option will terminate with probability 1. Assume that the agent is currently executing the hallway option and its current location is the cell adjacent to the target hallway, and also the door in the target hallway is locked. Then if the agent is holding the key, it continues executing the hallway option which will unlock the door and takes the agent to the target hallway, based on the stochastic process explained above. Once the agent exits the hallway, the door within that hallway changes its state to locked and closes again. The initiation set of the hallway option comprises all the states within the room plus the non-target hallway state leading into room. Note that for the cell next to the target hallway, the option can be initiated if either the door is open, or the door is closed and the agent is holding the key.

Figure 3 shows the states of the key process. Note that only at state six is the key ready to unlock doors (i.e. the agent is holding the key). There are eleven states defined for the key process and the agent can select one of three primitive actions, *get-key* that is defined over states $S_0$ through $S_5$, *key-nop* that is defined on all key states and *putback-key* that is defined only at state $S_6$. Primitive action *key-nop* has a stochastic effect in that the agent may drop the key with probability 3/10 once taken at state $S_6$ and with probability 7/10 will not change the state of the key process. If the agent drops the key, the key state transitions to state $S_7$. If the *key-nop* action is taken at states $S_0$ through $S_5$ and $S_7$ through $S_{10}$, it will not change the state of the key process. The agent will advance the state of the process when action *get-key* is executed, or will reset the state of the process to $S_0$ if action *putback-key* is executed. We also provide a multi-step *pickup-key* Markov option (on top of the *get-key* primitive action). *Pickup-key* option's policy $\pi$ advances key state (with probability 1) until the key is ready (state $S_6$). The termination condition $\beta(.)$ for *pickup-key* option is 1 for state $S_6$, and zero for rest of the states. Its initiation set comprises all of the key states except state $S_6$. Note that, if the agent drops the key, it will take more steps to pick it up (10 steps). This is to encourage the agent to put back the key, once it reaches state $S_6$.

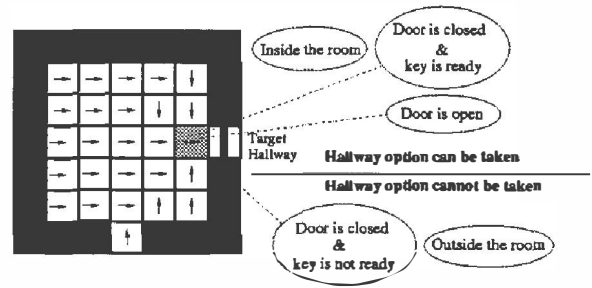

Figure 2: The policy associated with one of the hallway options. This figure also shows that the option can be taken at any cell within that room. In the shaded cell that is adjacent to the target hallway, the option can be taken if either the door is open, or the agent is holding the key at that cell. The option terminates in the target hallway and also in the shaded cell if the door is closed and the agent does is not holding the key.

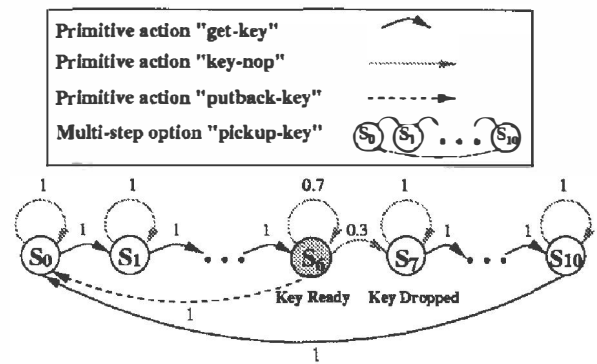

Figure 3: Representation of the key pickup option. At state $s_6$, agent is holding the key.

Figure 4 shows the evolution of the state of the environment when hallway options and key options run in parallel. Note that these options share the "key state" state variable, but they affect disjoint subspaces of the state space (e.g. hallway options will only control "position" and the "doors state" variables and key options will only affect the "key state" variable).

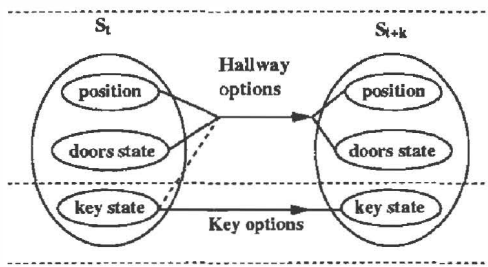

Figure 4: Evolution of states in the room navigation task. Initiation set, policy and termination probabilities of each option may share state variables with other parallel options (hallway options and pickup-key options share the "key state" variable), but each will affect mutually disjoint sets of state variables (hallway options control the position and the door state variables, and key option only controls the key state variable).

Using the notation developed in section 3, we define two classes of options: $C_1 = \{hallway_0, hallway_1, ..., hallway_7, room\_nop\}$[2] (eight hallway options and one *room-nop* option) and $C_2 = \{pickup\_key, key\_nop, putback\_key\}$[3]. For any hallway option, the set of state variables $W_{navigation} = \{position, doors\_state, key\_state\}$ (where *doors_state* specifies the state of each door) and for the key option $W_{key} = \{key\_state\}$. Note that $\Omega_{navigation} = \{position, doors\_state\}$ and $\Omega_{key} = \{key\_state\}$. Based on equation 8, the state space of the overall process is spanned by the set of state variables $W = W_{navigation} \cup W_{key} = \{position, doors\_state, key\_state\}$. The multi-options are members of the set $C_1 \times C_2$. Since only two hallway options can be taken in each room plus the *room-nop* option, and three key options can be taken in any cell within each room, a maximum of 9 multi-options can be defined in every state.

In order to evaluate the concurrent options framework, we compare its performance (for both definitions of the termination event) for the rooms example with the standard options framework. Note that in standard options framework, only one option can be taken at a time. Based on the concurrent options theorem in the previous section, we can apply SMDP-based Q-learning. Each multi-option is viewed as an indivisible, opaque unit of action (although intra-multi-option methods could also be developed). When multi-option $\vec{o}$ is initiated in state $s$, it transitions to state $s'$ in which multi-option $\vec{o}$ terminates according to the termination condition defined for multi-options. We can use SMDP Q-learning method (Bradtke & Duff, 1995; Sutton et al., 1999) to update the multi-option-value function $Q(s, \vec{o})$ after each decision epoch where the multi-option $\vec{o}$ is taken in some state $s$ and terminates in $s'$:

$$Q(s,\vec{o}) \leftarrow Q(s,\vec{o}) + \alpha \left[ r + \gamma^k \max_{\vec{o}' \in \vec{O}(s')} Q(s',\vec{o}') - Q(s,\vec{o}) \right]$$

where $k$ denotes the number of time steps since initiation of the multi-option $\vec{o}$ at state $s$ and its termination at state $s'$, and $r$ denotes the cumulative discounted reward over this period. For a fixed position as the starting point (upper left corner cell) and a fixed goal (hallway $H3$) in Figure 1, we used both frameworks in order to learn the policy to navigate from starting position to the target. For any primitive action, a reward of $-1$ is provided as the single step reward in order to learn a policy that optimizes the time for performing the task. Figure 5 shows the median of time (in terms of number of primitive actions taken until success) where for trial n, it is the median of all trials from 1 to n. This figure shows that the concurrent options framework based on the either of the termination events, learns a more optimal policy than the standard options framework. Moreover, the policy learned based on $T2$ is also more optimal than the policy learned by $T1$. One justification for this result is the SMDP process based on $T2$ allows less parallelism and hence less stochasticity during learning. This figure also shows that the standard options framework, converges faster compared to the concurrent options framework based on $T1$, but eventually, the concurrent options framework learns a better policy. Also in Figure 1, the small solid rectangles within cells represent the approximate location that the "pickup-key" option is initiated as the agent moves toward hallway $H0$ and the goal. With the policy learned using sequential execution of options, the agent navigates to the cell adjacent to the target hallway with the locked door using the corresponding hallway option, then it initiates the pickup-key option and waits in that cell for 6 steps until the key state advances to state $S_6$ at which point the key is ready to use. By overlapping execution of the hallway option and pick-up key option, the agent minimizes the time in order to reach the goal.

---

[2] Note that even though hallway options of different rooms also control disjoint sub-spaces of the state space, we put them in the same class since based on their initiation set, they can not run in parallel.

[3] Note that *room-nop*, *key-nop* and *putback-key* are primitive actions, but they are special case of single-step options.

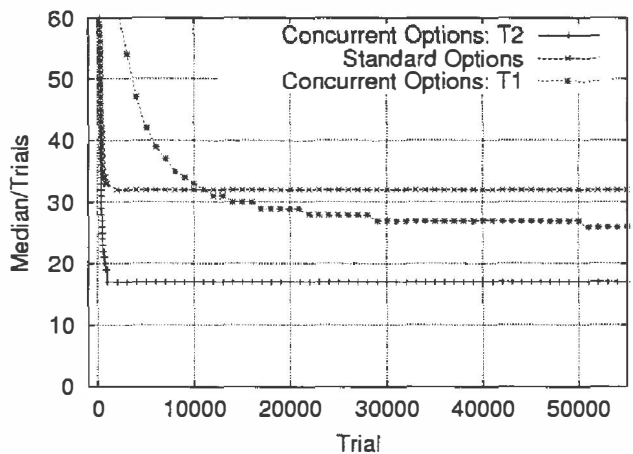

Figure 5: Performance of standard options and concurrent options framework using SMDP Q-learning in rooms example. The horizontal axis represents the number of trials, and the vertical axis shows the median of number of steps until success across trials. T1 and T2 denote the first and second definitions of the termination event, respectively.

## 5 Discussion and Future Work

In this paper we introduced the concurrent options model, which formalizes planning under uncertainty with parallel temporally extended actions where each action affects mutually disjoint subspaces of the environment state space. The key assumption that is required for a set of options to be safely run concurrently is that they are Markov. This restriction is necessary in order to have well defined termination condition and state prediction (transition probabilities). Consider the counter-example of a semi-Markov navigation option where the agent moves around the perimeter of a room twice before deciding to exit the room. For the key option to be invoked in parallel with this semi-Markov option, the agent would have to know how long the exit-room option had been running (information not available in the current decision epoch). We provided a simple computational experiment which shows that planning with concurrent options is more effective than when only one option is executed at a time. There is a clear connection between the model of concurrent options proposed in this paper to work on factored MDPs (Boutilier & Goldszmidt, 1995; Boutilier et al., to appear; Dean & Givan, 1997; Koller & Parr, 1999). Our approach relies on factoring the set of state variables using sets of behaviors that do not conflict. However, we do not use compact models of actions, such as dynamic Bayesian nets. One immediate problem for future research is to investigate how to represent options using DBNs, which would then facilitate compact representations of multi-options as well. Compact representations of options would also provide for a form of value function approximation, an issue that we have ignored in this paper. There are many interesting directions for further research: (1) Planning and learning with concurrent options when semi-Markov options are also included (in addition to Markov options). (2) Investigate learning of factored value functions for policies in multi-options (Koller & Parr, 1999). (3) Alleviate the *coherency* constraint introduced in section 3 to include cases when options can also modify shared state variables. (4) Finally, we should investigate the termination of a multi-option when interrupting a multi-option has higher value than continuing the multi-option (Sutton et al., 1999).